\documentclass[11pt]{article}

\usepackage[margin=1.25in, top=1in, bottom=1.1in]{geometry}
\usepackage[utf8]{inputenc}    % pdflatex: UTF-8 source encoding
\usepackage[T1]{fontenc}       % pdflatex: modern font encoding
\usepackage[autostyle=true]{csquotes}   % needed by some natbib styles; harmless otherwise
\usepackage{natbib}            % \citep, \citet
\usepackage{hyperref}
\usepackage{url}
\usepackage{booktabs}
\usepackage{amsmath}
\usepackage{amssymb}
\usepackage{amsfonts}
\usepackage{nicefrac}
\usepackage{microtype}
\usepackage{xcolor}
\usepackage{graphicx}
\usepackage{tikz}
\usepackage{algorithm}
\usepackage{algpseudocode}
\usepackage{multirow}
\usepackage{array}
\usepackage{tabularx}

\usetikzlibrary{arrows.meta, positioning, shapes.geometric, fit, backgrounds}

\newcommand{\blt}{\textsc{BLT}}
\newcommand{\bytekaz}{\textsc{ByteKaz}}
\newcommand{\matt}{\textsc{MATT}}
\newcommand{\llava}{\textsc{LLaVA}}
\newcommand{\reals}{\mathbb{R}}
\newcommand{\vocab}{\mathcal{V}}

\title{KazByte: Adapting Qwen models to Kazakh via Byte-level Adapter}

\author{%
  \textbf{Rauan Akylzhanov} \\
  \textit{Independent Researcher} \\
  \textit{Almaty, Kazakhstan} \\
  \url{https://ra312.github.io} \\
  \texttt{akylzhanov.r@gmail.com} \\
}

\begin{document}

\maketitle

% ─── Abstract ─────────────────────────────────────────────────────────────────
\begin{abstract}
Large language models fragment Kazakh text into many more tokens than
equivalent English text, because their tokenizers were built for
high-resource languages. This \emph{tokenizer tax} inflates compute,
shortens the effective context window, and weakens the model's grip on
Kazakh morphology.

We propose to bypass the tokenizer entirely by feeding raw bytes through a
small adapter that learns to speak the internal language of a frozen
Qwen2.5-7B. Once the adapter is trained, we freeze it and fine-tune only
the attention layers of Qwen on Kazakh text. Our \textbf{central hypothesis}
is that this two-stage process---first teach the interface, then adapt the
model---should \textbf{match or exceed} the accuracy of the original
Qwen2.5-7B on standard Kazakh benchmarks.

This report describes the \bytekaz{} architecture and training protocol.
Empirical validation is ongoing; this version stakes the design and
hypotheses for the record.
\end{abstract}

% ─── 1. Introduction ──────────────────────────────────────────────────────────
\section{Introduction}
\label{sec:intro}

Recent large language models (LLMs) have achieved remarkable multilingual
capability~\citep{grattafiori2024llama3, qwen2024qwen25}. Yet every pretrained
model is inextricably tied to a fixed tokenizer whose vocabulary is
determined before training. For lower-resource or morphologically complex
languages, this coupling creates persistent inefficiencies that neither
prompt engineering nor standard fine-tuning can fully resolve.

\paragraph{The Kazakh tokenizer problem.}
Kazakh is a Turkic, agglutinative language with rich suffixal morphology,
written primarily in Cyrillic with an ongoing transition to Latin script.
A single inflected verb form meaning \textit{``from your act of running''}
is a single semantic unit but is
tokenized into 10--12 BPE tokens by the Qwen tokenizer---roughly 5$\times$
the cost of a comparable English word.
This token fertility disparity has compounding consequences: longer
sequences for the same byte budget, fragmented subword units, and weaker
coverage of Kazakh-specific strings in the BPE vocabulary.

\paragraph{Why weight remapping does not work.}
A natural first instinct is to replace or extend the tokenizer and remap
Qwen's weights to the new representation. This is not feasible without
substantial retraining. A pretrained LLM learns (i) an embedding matrix
$E \in \reals^{|\vocab| \times d}$ tied to BPE vocabulary statistics,
(ii) token co-occurrence patterns baked into all feed-forward and attention
layers, and (iii) positional patterns calibrated to BPE token granularity
(\emph{ca.}\ 4 bytes/token for English). Swapping the input representation
shifts the input embedding distribution, invalidates positional statistics,
and misaligns all learned co-occurrence structure. This is a
\emph{network-wide distribution shift}, not a remapping problem.

\paragraph{Our proposal.}
We propose \bytekaz{}, illustrated in Figure~\ref{fig:architecture}, which
sidesteps the tokenizer at the boundary by learning a bidirectional byte
interface into Qwen2.5-7B. Patching follows the Byte Latent
Transformer~\citep{pagnoni2025blt}: entropy-based boundaries group bytes into
patches processed by a compact sequence of vectors through the global model.
The architecture is analogous to \llava~\citep{liu2024llava}: a modality
encoder maps non-token input into the LLM's space. Here the modality is
bytes; we add a decoder so both input and output are tokenizer-free.

\textbf{Hypothesis (staged training).} \textbf{(i)~Stage~A:} train the adapter
(encoder, projections, decoder) with \textbf{Qwen2.5-7B frozen}, so the
interface learns to present patch vectors Qwen can process. \textbf{(ii)~Stage~B:}
\textbf{freeze the adapter} and update \textbf{only attention-related weights}
in Qwen (per-layer $W_Q,W_K,W_V,W_O$ and, if desired, pre-attention LayerNorm;
\textbf{MLP blocks stay frozen}) on Kazakh continued pretraining or LM data.
The goal is to adapt attention to the new sequence geometry without
re-learning the full FFN capacity from scratch. \textbf{(iii)~Optional task
SFT} on Kazakh instruction or benchmark formats. We hypothesise Kazakh
accuracy should \textbf{match or exceed} Qwen2.5-7B with the stock BPE
tokenizer on the same evaluation suite---an empirical claim to be tested.

\paragraph{Contributions.}
\begin{itemize}
  \item \textbf{Architecture}: Bidirectional byte adapter (\blt{}-style local
        encoder + projections + Qwen2.5-7B body + local decoder).
  \item \textbf{Training hypothesis}: Stage~A (adapter only, Qwen frozen);
        Stage~B (adapter frozen, attention-only tuning in Qwen on Kazakh);
        optional task SFT.
  \item \textbf{Analysis}: Failure modes and mitigations (cold start, RoPE,
        masking, capacity, baseline comparison).
  \item \textbf{Data}: Verified list of openly available Kazakh corpora on
        Hugging Face (Section~\ref{sec:hf_kk_data}).
  \item \textbf{Evaluation plan}: SozKZ-style suite---MC QA
        (\texttt{kk-socio-cultural-bench-mc}), Belebele (\texttt{kaz\_Cyrl}),
        SIB-200 (\texttt{kk})~\citep{tukenov2026sozkz}; primary baselines:
        Qwen2.5-7B + BPE and SozKZ-600M~\citep{tukenov2026sozkz}.
\end{itemize}

% ─── 2. Related Work ──────────────────────────────────────────────────────────
\section{Related Work}
\label{sec:related}

\paragraph{Tokenizer transfer for low-resource languages.}
A line of work initialises new token embeddings as (weighted) averages of
existing ones. WECHSEL~\citep{minixhofer2022wechsel} uses cross-lingual
FastText similarities; FOCUS~\citep{dobler2023focus} exploits overlapping
vocabulary; Tik-to-Tok~\citep{remy2023tiktotok} and
Transtokenization~\citep{remy2024transtokenization} refine alignment via
translation dictionaries; TokAlign~\citep{li2025tokalign} uses GloVe
co-occurrence matrices. All of these remain within the BPE paradigm---new
tokens are still drawn from a fixed vocabulary---and do not address the
fundamental sequence-length penalty from high token fertility.

\matt{}~\citep{haltiuk2025matt} is the most recent and strongest entry:
it aligns attention patterns between a teacher (original tokenizer) and a
student (new tokenizer) via an Attention Influence Modelling (AIM)
objective, recovering large fractions of model quality with only a few GPU
hours. \bytekaz{} is complementary: \matt{} improves an extended BPE
tokenizer, while \bytekaz{} eliminates the tokenizer entirely.

\paragraph{Byte and character language models.}
ByT5~\citep{xue2022byt5} operates directly on UTF-8 bytes but uses an
encoder-decoder architecture trained from scratch, without a pretrained LLM
body. MegaByte~\citep{yu2023megabyte} introduces a hierarchical byte model
with a global and local transformer, but again trains end-to-end from
scratch. The Byte Latent Transformer~\citep{pagnoni2025blt} achieves
training-FLOP parity with Llama 3 by dynamically grouping bytes into
entropy-based patches; it is the direct architectural inspiration for
\bytekaz{}, but it does not leverage a pretrained LLM body.
H-Net~\citep{hwang2025hnet} proposes dynamic chunking for hierarchical
sequence modelling but is similarly trained from scratch.

\paragraph{Modality adapters for frozen LLMs.}
\llava~\citep{liu2024llava} connects a CLIP vision encoder to a frozen
LLaMA/Vicuna body via a linear projection, enabling image understanding
without retraining the core model. InstructBLIP~\citep{dai2023instructblip}
extends this with a Q-Former adapter. \bytekaz{} applies the same frozen-LLM
adapter paradigm to the byte modality, with the additional challenge of
requiring a \emph{decoder} adapter as well---both input and output must pass
through the byte interface.

\paragraph{Kazakh NLP.}
Dedicated Kazakh language resources remain sparse. KazNLP provides annotated
datasets for NER, NLI, and QA. FLORES-200 includes Kazakh as a low-resource
translation direction. To our knowledge, no prior work has addressed the
tokenizer fertility problem for Kazakh using architecture-level interventions.

% ─── 3. Background ────────────────────────────────────────────────────────────
\section{Background}
\label{sec:background}

\subsection{The BPE Tokenizer Tax on Agglutinative Languages}

Let $w$ be a word and $\mathrm{BPE}(w)$ its tokenization. Define the
\emph{token fertility} as $f(w) = |\mathrm{BPE}(w)|$. For English,
$\mathbb{E}[f(w)] \approx 1.3$; for Kazakh under the Qwen2.5 tokenizer,
$\mathbb{E}[f(w)]$ is typically much higher---a large disparity in sequence
length for the same text. Longer sequences increase training and inference
cost; exact factors depend on implementation (e.g.\ FlashAttention, KV cache).

\subsection{Byte Latent Transformer}

\citet{pagnoni2025blt} introduce the Byte Latent Transformer (\blt{}), which
replaces fixed-vocabulary tokenization with a dynamic, learnable byte-to-patch
mapping. Given a byte sequence $\mathbf{x} = (x_1, \ldots, x_n)$, a small
byte-level language model estimates the next-byte entropy
\begin{equation}
  H(x_i) = -\sum_{v \in \{0,\ldots,255\}} p_e(x_i = v \mid x_{<i}) \log p_e(x_i = v \mid x_{<i}).
  \label{eq:entropy}
\end{equation}
A new patch boundary is created whenever $H(x_i) > \theta_g$ for a global
threshold $\theta_g$. Low-entropy byte spans (predictable suffixes, common
word endings) are merged into long patches; high-entropy positions (e.g.\
word onsets, rare characters) receive dedicated patches.

The \blt{} architecture comprises three modules:
\begin{enumerate}
  \item \textbf{Local Encoder}: a lightweight transformer over the byte
        sequence that produces patch representations via cross-attention.
  \item \textbf{Global Latent Transformer}: a large transformer over the
        patch sequence---the primary locus of computation.
  \item \textbf{Local Decoder}: a lightweight transformer that predicts
        the next byte from patch context via cross-attention.
\end{enumerate}

% ─── 4. Method ────────────────────────────────────────────────────────────────
\section{The \bytekaz{} Architecture}
\label{sec:method}

\subsection{Overview}

\begin{figure}[t]
\centering
\begin{tikzpicture}[
  box/.style={draw, rounded corners=4pt, minimum width=6cm,
              minimum height=0.9cm, align=center, font=\small},
  trainbox/.style={box, fill=blue!10, draw=blue!60},
  frozenbox/.style={box, fill=gray!15, draw=gray!50},
  arrow/.style={-{Stealth[length=6pt]}, thick},
  label/.style={font=\footnotesize\itshape, text=gray}
]

\node[trainbox]  (enc)  at (0,  5.5) {\textbf{BLT Local Encoder} \\ \small 6--8 layers, $d_\ell{=}512$, $\approx$150M params};
\node[trainbox]  (penc) at (0,  4.0) {\textbf{Projection} $W_{\mathrm{enc}}$ \\ \small $512 \to 4096$, $\approx$2M params};
\node[frozenbox] (qwen) at (0,  2.5) {\textbf{Qwen2.5-7B Body} \\ \small frozen in Stage~A; attention tunable in Stage~B};
\node[trainbox]  (pdec) at (0,  1.0) {\textbf{Projection} $W_{\mathrm{dec}}$ \\ \small $4096 \to 512$, $\approx$2M params};
\node[trainbox]  (dec)  at (0, -0.5) {\textbf{BLT Local Decoder} \\ \small 6--8 layers, $d_\ell{=}512$, $\approx$150M params};

\draw[arrow] (0, 7.0) -- (enc)  node[midway, label, right=2pt]{raw bytes $(x_1,\ldots,x_n)$};
\draw[arrow] (enc)  -- (penc) node[midway, label, right=2pt]{patches $\in \reals^{m \times 512}$};
\draw[arrow] (penc) -- (qwen) node[midway, label, right=2pt]{projected patches $\in \reals^{m \times 4096}$};
\draw[arrow] (qwen) -- (pdec) node[midway, label, right=2pt]{hidden states $\in \reals^{m \times 4096}$};
\draw[arrow] (pdec) -- (dec)  node[midway, label, right=2pt]{patch context $\in \reals^{m \times 512}$};
\draw[arrow] (dec)  -- (0,-2.0) node[midway, label, right=2pt]{byte logits $\in \reals^{n \times 256}$};

\node[label, right=3.5cm] at (enc.east)  {\textcolor{blue!70}{trained}};
\node[label, right=3.5cm] at (penc.east) {\textcolor{blue!70}{trained}};
\node[label, right=3.5cm] at (qwen.east) {\textcolor{gray!70}{frozen}};
\node[label, right=3.5cm] at (pdec.east) {\textcolor{blue!70}{trained}};
\node[label, right=3.5cm] at (dec.east)  {\textcolor{blue!70}{trained}};

\end{tikzpicture}
\caption{The \bytekaz{} architecture. Blue modules are trained in Stage~A.
The Qwen2.5-7B stack is frozen during adapter training; in Stage~B the
adapter is frozen and attention sublayers in Qwen are updated on Kazakh data.
Bytes are grouped into entropy-based patches, projected into Qwen's space, and
processed by the transformer; the local decoder predicts bytes autoregressively.}
\label{fig:architecture}
\end{figure}
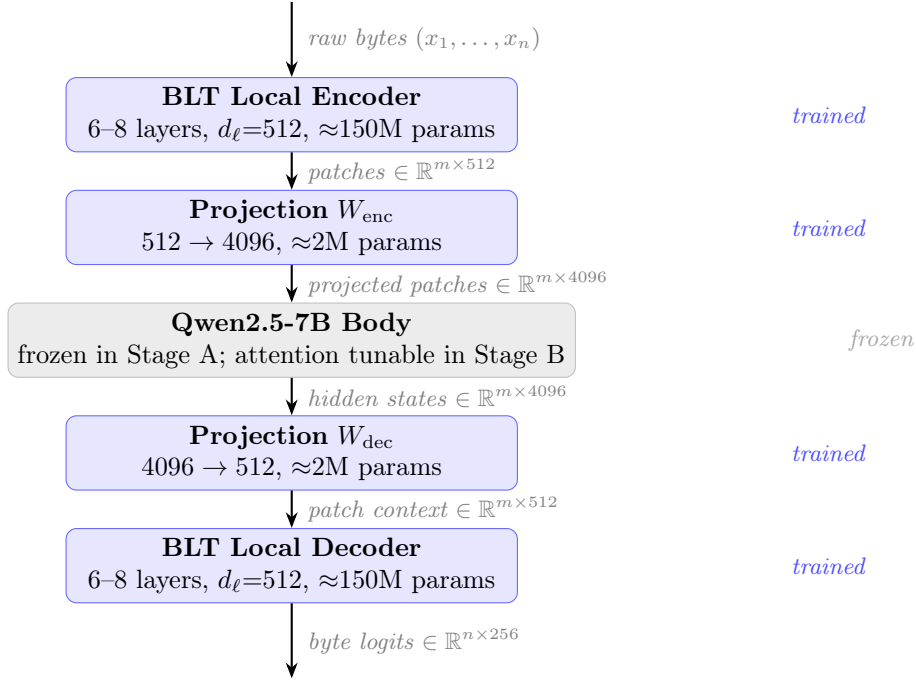

\bytekaz{} replaces Qwen2.5-7B's embedding table and LM head with learned byte
interfaces. During Stage~A the transformer body is frozen; Stage~B unfreezes
attention weights only (see Section~\ref{sec:training}).
Figure~\ref{fig:architecture} and Table~\ref{tab:components} summarise the
components.

\begin{table}[t]
\centering
\caption{Components of \bytekaz{}, their sizes, and training status.}
\label{tab:components}
\begin{tabular}{llrrr}
\toprule
Component & Role & Dim & Params & Status \\
\midrule
BLT Local Encoder   & Bytes $\to$ patches      & 512    & $\approx$150M & \textbf{Trained} \\
Projection $W_{\mathrm{enc}}$ & Patches $\to$ Qwen space & $512{\to}4096$ & $\approx$2M & \textbf{Trained} \\
Qwen2.5-7B body     & Sequence modelling       & 4096   & $\approx$7B   & Stage A: frozen; Stage B: attention trainable \\
Projection $W_{\mathrm{dec}}$ & Qwen space $\to$ patches & $4096{\to}512$ & $\approx$2M & \textbf{Trained} \\
BLT Local Decoder   & Patches $\to$ bytes      & 512    & $\approx$150M & \textbf{Trained} \\
\midrule
\multicolumn{3}{l}{Trainable total} & $\approx$304M & \\
\multicolumn{3}{l}{Frozen total}    & $\approx$7B   & \\
\bottomrule
\end{tabular}
\end{table}

\subsection{Local Encoder}

The local encoder is a causal transformer $f_\phi$ with $L_\ell = 6$--$8$
layers and hidden dimension $d_\ell = 512$. It ingests the raw byte sequence
$(x_1, \ldots, x_n)$---with bytes embedded via a learned table
$E_b \in \reals^{256 \times d_\ell}$---and produces per-byte hidden states.
Patch representations are then obtained via cross-attention over the byte
hidden states, with one query vector per patch boundary:
\begin{equation}
  \mathbf{p}_j = \mathrm{CrossAttn}\!\left(
    \mathbf{q}_j,\;
    \{h_i : x_i \in \mathrm{patch}_j\}
  \right),
  \quad j = 1,\ldots,m,
\end{equation}
where $\mathbf{q}_j \in \reals^{d_\ell}$ is a learned query for patch $j$
and $m \ll n$ is the (dynamic) number of patches.

Patch boundaries are determined by the entropy criterion of
Equation~\eqref{eq:entropy}, computed by a separate small byte LM
($\approx$10M parameters) pre-trained on the training corpus.

\subsection{Projection Layers}

Two linear projections bridge the encoder/decoder space and Qwen's space:
\begin{align}
  \tilde{\mathbf{p}}_j &= W_{\mathrm{enc}}\, \mathbf{p}_j + \mathbf{b}_{\mathrm{enc}},
  \quad W_{\mathrm{enc}} \in \reals^{4096 \times 512}, \label{eq:proj_enc} \\
  \hat{\mathbf{p}}_j   &= W_{\mathrm{dec}}\, \mathbf{h}_j  + \mathbf{b}_{\mathrm{dec}},
  \quad W_{\mathrm{dec}} \in \reals^{512  \times 4096}. \label{eq:proj_dec}
\end{align}
Both projections are followed by LayerNorm. To avoid cold-start instability
(Section~\ref{sec:discussion}), $W_{\mathrm{enc}}$ is initialised so that
$\tilde{\mathbf{p}}_j$ matches Qwen's embedding statistics:
\begin{equation}
  W_{\mathrm{enc}} \sim \mathcal{N}\!\left(0,\; \frac{\sigma_{\mathrm{emb}}^2}{d_\ell}\right),
  \label{eq:init}
\end{equation}
where $\sigma_{\mathrm{emb}}^2$ is the empirical variance of Qwen's embedding
matrix.

\subsection{Qwen2.5-7B Body}

Qwen2.5-7B's transformer stack receives the projected patch sequence
$(\tilde{\mathbf{p}}_1, \ldots, \tilde{\mathbf{p}}_m)$ with a standard
causal attention mask. We remove the embedding table and LM head entirely;
only the attention and feed-forward layers are invoked. Rotary position
embeddings (RoPE) use patch indices as position values, which is valid
because RoPE is index-based rather than fixed-length.

\subsection{Local Decoder}

The local decoder is a causal transformer $g_\psi$ symmetric to the encoder
($L_\ell = 6$--$8$ layers, $d_\ell = 512$). For each patch $j$, it autore-
gressively predicts the bytes of $\mathrm{patch}_j$ conditioned on the patch
context $\hat{\mathbf{p}}_j$ from Equation~\eqref{eq:proj_dec} and all
previously generated bytes via cross-attention:
\begin{equation}
  p(x_i \mid x_{<i},\; \hat{\mathbf{p}}_{1:j}) =
    \mathrm{softmax}\!\left(
      g_\psi\!\left(x_{<i},\; \hat{\mathbf{p}}_{1:j}\right)_i \cdot E_b^\top
    \right).
\end{equation}
Causal masking ensures that bytes in patch $j$ attend only to Qwen context
from patches $< j$ and to preceding bytes within the same patch.

\subsection{Analogy to LLaVA}

\bytekaz{} is structurally analogous to \llava~\citep{liu2024llava}: both
use a small modality-specific encoder, a linear projection, and a frozen LLM
body. Table~\ref{tab:llava_comparison} highlights the key differences.

\begin{table}[t]
\centering
\caption{Comparison of \llava{} and \bytekaz{}.}
\label{tab:llava_comparison}
\begin{tabular}{lll}
\toprule
 & \llava & \bytekaz \\
\midrule
Input modality          & Image patches (CLIP)            & Byte patches (BLT encoder) \\
Output modality         & Text tokens (BPE LM head)       & Byte sequence (BLT decoder) \\
Frozen body             & LLaMA / Vicuna                  & Qwen2.5-7B (Stage A) \\
Trainable components    & CLIP encoder + projection       & Encoder + decoder + projections \\
Bidirectional interface & Input only                      & Input \textbf{and} output \\
Target application      & Vision-language tasks           & Agglutinative language NLP \\
\bottomrule
\end{tabular}
\end{table}

% ─── 5. Training Protocol ─────────────────────────────────────────────────────
\section{Training Protocol}
\label{sec:training}

We adopt a curriculum: \textbf{Stage~A} (adapter alignment, Qwen frozen),
\textbf{Stage~B} (adapter frozen, attention-only tuning on Kazakh), optional
\textbf{Stage~C} (task SFT).

\subsection{Stage A: Representation Alignment (adapter trained, Qwen frozen)}

\textbf{Objective.} Train encoder, decoder, and projections; keep all Qwen2.5-7B
parameters frozen.

\textbf{Data.} English and Chinese text (1--2B tokens) so the frozen body sees
inputs near its pretraining distribution.

\textbf{Loss.} Primary term: byte-level cross-entropy through the adapter.
Optional \textbf{hidden-state alignment} (well-defined across byte vs.\ BPE):
on aligned text pairs, minimise MSE between patch-level hidden states from the
adapter path and segment-aggregated teacher states from a frozen Qwen2.5-7B
forward pass on BPE-tokenized text, in the spirit of attention / segment
alignment in \matt{}~\citep{haltiuk2025matt} rather than KL on mismatched
token vs.\ byte vocabularies:
\begin{equation}
  \mathcal{L} = \mathcal{L}_{\mathrm{CE}}(x) + \alpha \sum_{\ell \in \mathcal{S}}
    \bigl\| h^{(\ell)}_{\mathrm{patch}} - \operatorname{stopgrad}\bigl(h^{(\ell)}_{\mathrm{teacher}}\bigr) \bigr\|_2^2 .
  \label{eq:loss}
\end{equation}
Here $h^{(\ell)}_{\mathrm{patch}}$ are hidden states at selected layers
$\mathcal{S}$ along the patch path; $h^{(\ell)}_{\mathrm{teacher}}$ are
teacher states after pooling to patch boundaries; $\mathrm{sg}$ is stop-gradient.
Setting $\alpha{=}0$ is a valid baseline (CE only).

\textbf{Success signal.} BPB on held-out English/Chinese in a reasonable range;
optional alignment loss decreasing.

\subsection{Stage B: Kazakh Adaptation (adapter frozen, attention-only in Qwen)}

\textbf{Objective.} Freeze the adapter; train \textbf{only attention weights}
in Qwen2.5-7B (e.g.\ $W_Q,W_K,W_V,W_O$ per layer; optionally input LayerNorm
before attention). \textbf{MLP / FFN blocks remain frozen.} Rationale:
attention routes information between positions; the new patch geometry may
primarily require reweighting dependencies, while FFN stores much factual
capacity we wish to preserve.

\textbf{Data.} Kazakh text from the SozKZ pretraining corpus~%
\citep{tukenov2026sozkz,sozkz_corpus}: 9B tokens collected from 18 public
sources including CulturaX, HPLT~2.0, mC4, MADLAD-400, CC-100, Kazakh
Wikipedia, and others, cleaned via a 9-stage pipeline (48.2\% pass rate from
28.4M raw documents). Available on Hugging Face as
\texttt{saken-tukenov/sozkz-corpus-clean-v3}. Target: 5--10B tokens where
feasible; see also open corpora in Section~\ref{sec:hf_kk_data}.

\textbf{Patch-size calibration.} Ablate entropy threshold $\theta_g$; report
average patch size and BPB.

\subsection{Stage C: Task Fine-tuning (optional)}

\textbf{Objective.} Instruction following or benchmark-aligned formats (e.g.\
MC QA, reading comprehension). Adapter can remain frozen; continue
attention-only updates or add a small task head. Data: Kazakh instruction
sets, 50--200M tokens typical.

\begin{table}[t]
\centering
\caption{Training stages and resource estimates (indicative).}
\label{tab:schedule}
\begin{tabular}{llll}
\toprule
Stage & Tokens & Hardware & Est. Time \\
\midrule
A --- Adapter alignment (Qwen frozen) & 1--2B  & 4$\times$ A100 80\,GB & 1--2 weeks \\
B --- Kazakh LM, attention-only Qwen  & 5--10B & 4$\times$ A100 80\,GB & 2--4 weeks \\
C --- Task SFT                        & 50--200M & 2$\times$ A100 80\,GB & 2--3 days \\
\bottomrule
\end{tabular}
\end{table}

% ─── 7. Evaluation Plan ───────────────────────────────────────────────────────
\section{Evaluation Plan}
\label{sec:eval}

\subsection{Intrinsic Metrics}

\begin{itemize}
  \item \textbf{Bits-per-byte (BPB)} on held-out Kazakh text: primary
        language modelling quality metric.
  \item \textbf{Average patch size}: measures compression efficiency;
        higher is better.
  \item \textbf{Alignment loss} (optional MSE in Equation~\eqref{eq:loss}):
        how well patch-path hidden states match the teacher.
\end{itemize}

\subsection{Downstream Benchmarks}
\label{sec:downstream_benchmarks}

We evaluate on the same three Kazakh benchmarks as \citet{tukenov2026sozkz}
(Sect.~4.1 of arXiv:2603.20854). The following subsections follow that
paper's breakdown; Table~\ref{tab:benchmarks} summarises the suite.

\begin{table}[t]
\centering
\caption{Kazakh NLP benchmarks (SozKZ suite~\citep{tukenov2026sozkz}).}
\label{tab:benchmarks}
\small
\begin{tabular}{llll}
\toprule
Benchmark & Task & Metric & Random \\
\midrule
MC QA~\citep{kk_socio_cultural_bench} & 4-choice QA (7{,}111 items) & Accuracy & 25\% \\
Belebele~\citep{bandarkar2023belebele} & Reading comprehension & Accuracy & 25\% \\
SIB-200~\citep{adelani2024sib200}      & Topic classification (7-class) & Accuracy & 14.3\% \\
\bottomrule
\end{tabular}
\end{table}

\subsubsection{MC QA}

Multiple-choice cultural QA from \texttt{stukenov/kk-socio-cultural-bench-mc}~%
\citep{kk_socio_cultural_bench,tukenov2026sozkz}: 7{,}111 questions across
18 categories (Kazakh culture, history, traditions), four options each.
\textbf{Metric:} accuracy; random baseline 25\%. \textbf{Scoring:} full
answer-string likelihood (sum of token log-probabilities, length-normalised),
not single-token logit comparison, to avoid tokenizer-vocabulary bias.

\subsubsection{Belebele}

Reading comprehension from \texttt{facebook/belebele}~\citep{bandarkar2023belebele}
(subset \texttt{kaz\_Cyrl}): passage, question, four multiple-choice answers.
\textbf{Metric:} accuracy; random baseline 25\%. Same length-normalised
answer likelihood scoring as MC QA.

\subsubsection{SIB-200}

Topic classification from \texttt{Davlan/sib200}~\citep{adelani2024sib200}
(language \texttt{kk}): seven topic categories.
\textbf{Metric:} accuracy; random baseline 14.3\%. \textbf{Scoring:}
logit-based classification with Kazakh topic labels, following
\citet{tukenov2026sozkz}.

\subsection{Evaluation protocol}
\label{sec:eval_protocol}

To match the reproducibility style of \citet{tukenov2026sozkz} (Sect.~4.3):

\begin{itemize}
  \item \textbf{Logit-based / likelihood scoring.} Tasks are scored without
        text generation where possible, reducing sensitivity to decoding
        hyperparameters.
  \item \textbf{Multiple-choice tasks.} Each candidate answer is scored by the
        sum of its token log-probabilities conditioned on the prompt,
        normalised by token count (full answer likelihood).
  \item \textbf{Zero-shot.} No in-context exemplars and no task-specific
        fine-tuning before benchmark scoring, unless an ablation explicitly
        targets supervised adaptation.
  \item \textbf{Reporting.} Compare \bytekaz{} checkpoints (Stage~A/B) to
        the baselines listed in Section~\ref{sec:baselines}; primary reference
        is Qwen2.5-7B + BPE and SozKZ-600M~\citep{tukenov2026sozkz}.
\end{itemize}

\subsection{Baselines}
\label{sec:baselines}

\subsubsection{Multilingual and extended-tokenizer references}

\begin{enumerate}
  \item \textbf{Qwen2.5-7B + BPE} (primary baseline): stock tokenizer.
  \item \textbf{SozKZ-600M}~\citep{tukenov2026sozkz}: Llama-architecture model
        trained from scratch on $\sim$9B Kazakh tokens with a dedicated 50K BPE
        tokenizer; strongest open dedicated-Kazakh reference in the SozKZ
        evaluation suite.
  \item \textbf{Qwen2.5-7B + vocabulary extension}: extended Kazakh-aware
        BPE; continual pre-training.
\end{enumerate}

\subsubsection{Byte-level and \bytekaz{} checkpoints}

\begin{enumerate}
  \setcounter{enumi}{3}
  \item \textbf{BLT-7B fine-tuned on Kazakh}: byte LM without frozen Qwen body.
  \item \textbf{\bytekaz{} after Stage A} (adapter only; Qwen frozen).
  \item \textbf{\bytekaz{} after Stage B} (adapter frozen; attention-tuned Qwen).
\end{enumerate}

\subsection{Ablations}

\begin{enumerate}
  \item Patching strategy: entropy vs.\ fixed-stride vs.\ whitespace.
  \item Encoder size: $d_\ell \in \{512, 768, 1024\}$.
  \item Alignment weight: $\alpha \in \{0, 0.1, 0.5, 1.0\}$ in
        Equation~\eqref{eq:loss}.
  \item Attention-only vs.\ attention + LayerNorm vs.\ last-$k$ full layers
        (if needed) in Stage B.
  \item Script: Cyrillic Kazakh vs.\ Latin Kazakh; cosine distance between
        paired representations.
\end{enumerate}

% ─── Open Kazakh corpora (Hugging Face) ──────────────────────────────────────
\section{Open Kazakh Text Corpora on Hugging Face}
\label{sec:hf_kk_data}

We list \textbf{openly accessible} dataset repos on Hugging Face that include
a Kazakh (\texttt{kk}) subset or Kazakh-specific configuration, with
verification notes from the dataset cards (as of the writing of this note).
\textbf{Gated} datasets require login / acceptance; they are not fully open
without that step.

\begin{table}[t]
\centering
\caption{Verified Kazakh-capable corpora on Hugging Face (open access).}
\label{tab:hf_kk}
\small
\begin{tabular}{p{3.2cm}p{4.2cm}p{5.5cm}}
\toprule
Dataset ID & Kazakh indicator & License / access \\
\midrule
\texttt{statmt/cc100} & Config \texttt{kk}: $\approx$889M chars listed
  & No IP claim on corpus prep.; bound by Common Crawl ToU. \\
\texttt{wikimedia/wikipedia} & Config e.g.\ \texttt{20231101.kk}
  & GFDL + CC BY-SA 3.0 (Wikimedia dumps legal page). \\
\texttt{HPLT/HPLT2.0\_cleaned} & Subset \texttt{kaz\_Cyrl}; card lists
  \texttt{cc0-1.0}; not gated & Prefer \texttt{HPLT/HPLT3.0} if starting fresh. \\
\texttt{kurumikz/Cleaned-Kazakh-Wikipedia} & 228{,}810 articles; Cyrillic
  & ODC-BY (dataset card). \\
\texttt{legacy-datasets/mc4} & Language \texttt{kk} listed; \textbf{deprecated}
  in favor of \texttt{allenai/c4}
  & ODC-BY; use \texttt{allenai/c4} for new work. \\
\bottomrule
\end{tabular}
\end{table}

\paragraph{Gated or restricted (not fully open without acceptance).}
\texttt{oscar-corpus/OSCAR-2301} is \textbf{manually gated}; the card has
stated access suspension periods---check current status before relying on it.
\texttt{uonlp/CulturaX} is gated (accept conditions); includes Kazakh
(\texttt{kk}) in the language table; license follows mC4 and OSCAR per the
card.

% ─── 8. Expected Results and Discussion ───────────────────────────────────────
\section{Expected Results and Discussion}
\label{sec:discussion}

\paragraph{Sequence efficiency (hypothesis).}
Patch sequences may be shorter than BPE token sequences for Kazakh; we will
report measured patch counts, tokens per byte, and throughput separately from
downstream scores.

\begin{table}[t]
\centering
\caption{Reference accuracies (\%) from \citet{tukenov2026sozkz} on the
SozKZ suite; \bytekaz{} and Qwen2.5-7B rows to be measured.}
\label{tab:sozkz_reference}
\begin{tabular}{lccc}
\toprule
Method & MC QA & Belebele & SIB-200 \\
\midrule
SozKZ-600M (published) & 30.3 & 27.0 & 25.5 \\
Qwen-2.5-0.5B (published) & 31.5 & 30.0 & 19.1 \\
Qwen2.5-7B + BPE & --- & --- & --- \\
\bytekaz{} after Stage B & --- & --- & --- \\
\bottomrule
\end{tabular}
\end{table}

\paragraph{Benchmark quality (hypothesis).}
After Stage~B (attention-only on Kazakh), we hypothesise \bytekaz{} can
\textbf{match or exceed} Qwen2.5-7B + BPE on Kazakh tasks in
Table~\ref{tab:benchmarks}, because the pretrained Qwen trunk is retained
and only attention is adapted to the patch interface. BLT-7B without Qwen
may trail on knowledge-heavy tasks until heavily trained.

\paragraph{Alignment benefit.}
Hidden-state alignment (Equation~\eqref{eq:loss}) is optional; \matt{}-style
signals~\citep{haltiuk2025matt} may accelerate Stage~A stability---an
experiment.

% ─── 9. Conclusion ────────────────────────────────────────────────────────────
\section{Conclusion}
\label{sec:conclusion}

We have proposed \bytekaz{}, a byte-level adapter around Qwen2.5-7B with
\blt-style local encoder and decoder. The central hypothesis is
\textbf{staged training}: align the adapter with Qwen frozen (Stage~A), then
freeze the adapter and \textbf{tune only attention layers} on Kazakh (Stage~B),
with optional task SFT (Stage~C). We hypothesise Kazakh accuracy can
\textbf{match or exceed} the baseline Qwen2.5-7B + BPE tokenizer on the same
tasks---an empirical question.

We summarised failure modes, evaluation against Qwen2.5-7B + BPE, and a
verified list of open Kazakh corpora on Hugging Face
(Section~\ref{sec:hf_kk_data}). This note stakes the architecture and
protocol for collaborators; experiments are future work.

% ─── Limitations ──────────────────────────────────────────────────────────────
\section*{Limitations}

\textbf{Untested at scale.} This is a research proposal; no empirical
results are presented. All efficiency and quality estimates are projections
based on BLT and LLaVA analogues.

\textbf{Alignment objective.} Equation~\eqref{eq:loss} uses hidden-state MSE
optional; aligning teacher/student segments requires careful string alignment
implementation.

\textbf{Entropy model.} The byte LM for patch boundaries may be biased
toward high-resource languages; Kazakh-specific entropy models are optional.

\textbf{Qwen Kazakh exposure.} Stage~B (attention-only on Kazakh) is intended
to close the gap; if insufficient, selective FFN unfreezing is a fallback
ablation.

% ─── References ───────────────────────────────────────────────────────────────

\end{document}